\documentclass[final]{cvpr}

\usepackage{times}
\usepackage{epsfig}
\usepackage{graphicx}
\usepackage{amsmath}
\usepackage{amssymb}
\usepackage{enumitem}
\usepackage{multirow}
\usepackage{pifont}
\usepackage{amstext}
\setlist[itemize]{leftmargin=*}
\setlist[enumerate]{leftmargin=*}

\usepackage[pagebackref=true,breaklinks=true,colorlinks,bookmarks=false]{hyperref}

\newcommand{\correct}{\ding{51}}
\newcommand{\wrong}{\ding{55}}

\def\cvprPaperID{3090} 
\def\confYear{CVPR 2021}

\begin{document}

\title{POSEFusion: Pose-guided Selective Fusion for\\ Single-view Human Volumetric Capture}

\author{Zhe Li$^1$, Tao Yu$^1$, Zerong Zheng$^1$, Kaiwen Guo$^2$, Yebin Liu$^1$\\
$^1$Department of Automation, Tsinghua University, China\quad $^2$Google, Switzerland}

\maketitle


\begin{abstract}
    We propose \textbf{PO}se-guided \textbf{SE}lective \textbf{Fusion} (POSEFusion), a single-view human volumetric capture method that leverages tracking-based methods and tracking-free inference to achieve high-fidelity and dynamic 3D reconstruction. 
    By contributing a novel reconstruction framework which contains pose-guided keyframe selection and robust implicit surface fusion, our method fully utilizes the advantages of both tracking-based methods and tracking-free inference methods, and finally enables the high-fidelity reconstruction of dynamic surface details even in the invisible regions. 
    We formulate the keyframe selection as a dynamic programming problem to guarantee the temporal continuity of the reconstructed sequence. 
    Moreover, the novel robust implicit surface fusion involves an adaptive blending weight to preserve high-fidelity surface details and an automatic collision handling method to deal with the potential self-collisions. 
    Overall, our method enables high-fidelity and dynamic capture in both visible and invisible regions from a single RGBD camera, and the results and experiments show that our method outperforms state-of-the-art methods. 
\end{abstract}

\section{Introduction}
\label{sec:intro}
Human volumetric capture, due to their potential value in holographic communication, online education, games and the movie industry has been a popular topic in computer vision and graphics for decades. Multi-view camera array methods \cite{bradley2008markerless, gall2009motion, liu2009point, brox2009combined, liu2011markerless, ye2012performance, mustafa2015general, dou2016fusion4d, pons2017clothcap, leroy2017multi} can achieve high-fidelity human volumetric capture using multiple RGB or depth sensors but suffer from sophisticated equipment or run-time inefficiency, which limits their application deployment. In contrast, single-view human volumetric capture \cite{li2009robust, zollhofer2014real, guo2015robust, newcombe2015dynamicfusion, yu2018doublefusion, habermann2019livecap, saito2019pifu, saito2020pifuhd, habermann2020deepcap, li2020monocular, su2020robustfusion} has attracted more and more attention for its convenient setup.

\begin{figure}[t]
    \centering
    \includegraphics[width=\linewidth]{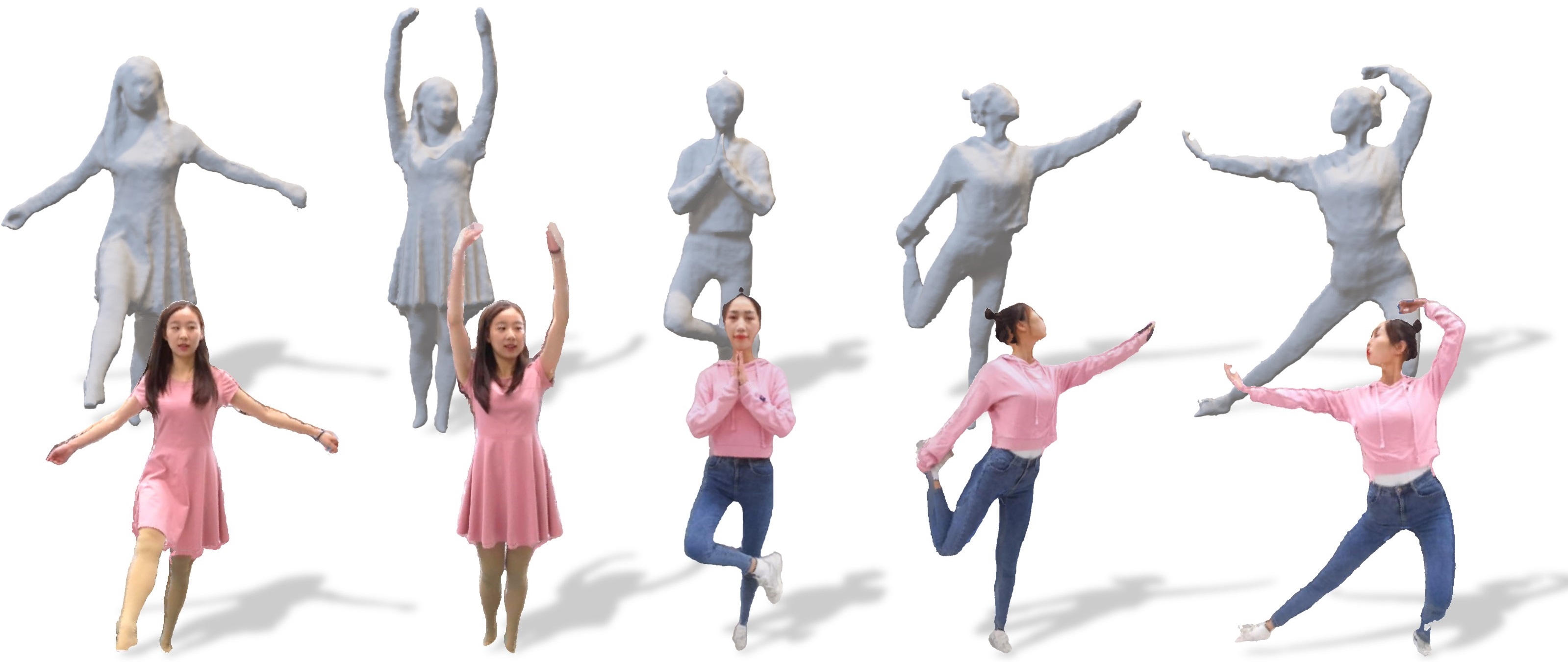}
    \caption{High-fidelity and dynamic results reconstructed using our method.}
    \label{fig:teaser}
    \vspace{-0.4cm}
\end{figure}

Current methods for single-view human volumetric capture can be roughly classified into two categories: tracking-based methods and tracking-free ones. Tracking-based methods utilize a pre-scanned template \cite{li2009robust, guo2015robust, habermann2019livecap, habermann2020deepcap} or continuously fused mesh \cite{newcombe2015dynamicfusion, yu2018doublefusion, su2020robustfusion} as the reference model, and solve or infer the deformation of the reference model parameterized by embedded skeletons \cite{ye2014real, bogo2015detailed, xu2018monoperfcap}, node graph \cite{li2009robust, newcombe2015dynamicfusion}, parametric body models (e.g., SMPL \cite{loper2015smpl}) \cite{zhi2020texmesh}, or a combination of them \cite{yu2017BodyFusion, yu2018doublefusion, habermann2019livecap, habermann2020deepcap}.
In these tracking-based methods, previous observations are integrated into the current frame after calculating the deformations across frames, thus plausible geometric details are preserved in the invisible regions. In addition, the reconstructed models are temporally continuous thanks to the frame-by-frame tracking. However, none of their deformation representations, neither skeletons nor node graph, is able to describe topological changes or track extremely large non-rigid deformations (Fig.~\ref{fig:qualitative comparison}(c)), which is an 
inherent drawback of the tracking-based methods.

On the other end of the spectrum, tracking-free methods \cite{varol2018bodynet, natsume2019siclope, gabeur2019moulding, smith2019facsimile, zhu2019detailed, alldieck2019tex2shape, zheng2019deephuman, saito2019pifu} mainly focus on geometric and/or texture recovery from a single RGB(D) image. 
By learning from a large amount of 3D human data, these methods demonstrate promising human reconstruction with high-fidelity details in visible regions and plausible shape in invisible areas \cite{saito2019pifu, saito2020pifuhd}. As the reconstruction for the current frame is independent from the previous frames, these methods can easily handle topological changes. However, their results may deteriorate in the cases of challenging human poses and/or severe self-occlusions. 
Besides, the reconstructions in the invisible regions are usually oversmoothed for the lack of observations (Fig.~\ref{fig:qualitative comparison}(d)). What's worse, tracking-free methods are incapable of generating temporally continuous results when applied on video inputs. 

By reviewing the advantages and the drawbacks of these two types of methods, it is easy to notice that tracking-based methods and tracking-free inference are naturally complementary as shown in Tab.~\ref{tab: comparison of methods}. 
A straightforward way is to combine both branches by integrating the inferred models of all the other frames in the monocular RGBD sequence into the current frame to recover the invisible regions. 
The benefits of such a pipeline are: a) topological changes and large deformations can be accurately reconstructed using tracking-free inference directly, b) the invisible surfaces can be faithfully recovered by integrating the other frames into current frame, and finally c) temporal continuity is guaranteed by tracking the whole sequence frame-by-frame. 
However, the afore-mentioned pipeline still has limitations. 
Specifically, if we fuse all the other frames indiscriminately, we can only generate static surfaces with all the dynamic changing details averaged together. Moreover, it remains difficult for such a pipeline to handle the artifacts caused by self-collisions. 
To this end, we further propose {PO}se-guided {SE}lective Fusion (POSEFusion), a novel pipeline that contains pose-guided keyframe selection and adaptive implicit surface fusion. In this pipeline, we only integrate the keyframes selected by our proposed pose-guided metric, which takes into account both visibility complementarity and pose similarity.

Our key observations are: a) keyframes with similar poses to the current frame enable the recovery of physically plausible dynamic details, b) keyframes with complementary viewpoints to the current frame avoid to oversmooth the visible regions, and c) the adaptive fusion considers depth observations and visibility promotes to preserve the surface details and resolves collision artifacts. Based on these observations, the limitations of the simple pipeline are successfully overcome. 

Specifically, we start with SMPL \cite{loper2015smpl} tracking for all the frames given a monocular RGBD sequence as input. We then utilize the SMPL model as a robust and effective proxy and propose a novel criterion to quantify pose similarity and visibility complementarity. Based on this criterion, we select appropriate keyframes for each frame. Note that per-frame keyframe selection cannot guarantee the temporal continuity of invisible details; therefore, we further formulate the selection as a dynamic programming of min-cost path to reconstruct dynamic and temporally continuous invisible details. In implicit surface fusion, we propose an adaptive blending weight which considers depth and visibility information to avoid oversmooth fusion and preserve the observed details. Finally, we propose an automatic collision handling scheme to deal with possible self-collisions while maintaining adjacent details. 

In summary, this paper proposes the following technical contributions:
\begin{itemize}
\setlength{\itemsep}{0pt}
\setlength{\parsep}{0pt}
\setlength{\parskip}{0pt}
\vspace{-0.2cm}
	\item A new human volumetric capture pipeline that leverages tracking-based methods and tracking-free inference, and achieves high-fidelity and dynamic reconstruction in both visible and invisible regions from a single RGBD camera (Sec.~\ref{sec:pipeline}).
	\item A new pose-guided keyframe selection scheme that considers both pose similarity and visibility complementarity and enables detailed and pose-guided reconstruction in the invisible regions (Sec.~\ref{sec:keyframe selection}).
	\item A robust implicit surface fusion scheme that involves an adaptive blending weight conditioned by depth observations and visibility, and an automatic collision handling method which considers an adjacent no-collision model into the fusion procedure to maintain the adjacent details while eliminating collision artifacts (Sec.~\ref{sec:implicit surface fusion}).
\vspace{-0.2cm}
\end{itemize}
Building on these novel techniques, POSEFusion is the first single-view approach that is able to capture high-fidelity and dynamic details in both visible and invisible regions.
Given a monocular RGBD sequence as input, our method is able to produce compelling human reconstruction results with complete, dynamic, temporally continuous, and high-fidelity details. 
The experimental results prove that our method outperforms state-of-the-art methods. 

\begin{table*}[t]
    \footnotesize
    \centering
    \begin{tabular}{ccccccc}
      \hline
      \multicolumn{2}{c}{Methods}                                                                      & \begin{tabular}[c]{@{}c@{}}Topological\\ Change\end{tabular} & \begin{tabular}[c]{@{}c@{}}Natural\\ Deformation\end{tabular} & \begin{tabular}[c]{@{}c@{}}Details in\\ Invisible Regions\end{tabular}    & \begin{tabular}[c]{@{}c@{}}Temporal\\ Continuity\end{tabular} & Texture\\
      \hline
      \textbf{Tracking-based} & DoubleFusion \cite{yu2018doublefusion} & \wrong & \wrong & {\correct (Static)} & \correct & None\\
      \cline{2-7}
      & RobustFusion \cite{su2020robustfusion} & \wrong & \wrong & {\correct (Static)} & \correct & {\correct (Low-quality)}\\
      \cline{2-7}
      & TexMesh \cite{zhi2020texmesh} & \wrong & \wrong & \correct (Low-quality) & \correct & \correct (Static)\\
      \hline
      \textbf{Tracking-free} & PIFu \cite{saito2019pifu}/PIFuHD \cite{saito2020pifuhd} & \correct & \correct & \wrong & \wrong & {\correct (Low-quality)}\\
      \hline
      \multicolumn{2}{c}{\textbf{Ours}} & \correct & \correct & \begin{tabular}[c]{@{}c@{}}\ding{51}\\ (Dynamic, High-quality)\end{tabular} & \correct & \begin{tabular}[c]{@{}c@{}}\ding{51}\\ (Dynamic, High-quality)\end{tabular}\\
      \hline
    \end{tabular}
    \caption{Comparison of our method with other state-of-the-art works when applying a monocular RGBD video as input. Our method inherits all the advantages of tracking-based and tracking-free methods while avoiding their drawbacks. Moreover, our method can reconstruct dynamic pose-guided geometric details in both visible and invisible regions.}
    \label{tab: comparison of methods}
    \vspace{-0.4cm}
\end{table*}

\section{Related Work}

\subsection{Tracking-based Human Reconstruction}
Some works in tracking-based methods utilize a pre-scanned person-specific model as a template, and deform it to fit with depth input of each frame. Especially, for human reconstruction, Gall \textit{et al.} \cite{gall2009motion} and Liu \textit{et al.} \cite{liu2011markerless} modeled body motion by skeletons embedded in the template. Besides skeletal motion, embedded deformation graph \cite{Sumner2007embedded} is an alternative parameterization method for non-rigid reconstruction. Li \textit{et al.} \cite{li2009robust} solved the warp field modeled by \cite{Sumner2007embedded} and reconstructed detailed 3D geometric sequences from a single-view depth stream. Zollh{\"o}fer \textit{et al.} \cite{zollhofer2014real} enabled real-time performance for general non-rigid tracking based on the parallelism of GPU. Guo \textit{et al.} \cite{guo2015robust} introduced a $L_0$-based regularizer to implicitly constrain articulated motion. LiveCap \cite{habermann2019livecap} utilized a person-specific template and achieved real-time monocular performance capture. On another branch, volumetric fusion methods replaced the pre-scanned template with a continuously fused model for online incremental reconstruction. The pioneering work KinecFusion \cite{izadi2011kinectfusion} reconstructed a rigid scene incrementally by using a commercial RGBD camera. The following work \cite{niessner2013real, dai2017bundlefusion, lee2020texturefusion} focused on memory cost, geometry, and texture quality for rigid scene reconstruction, respectively. DynamicFusion \cite{newcombe2015dynamicfusion} extended \cite{izadi2011kinectfusion} and introduced a dense non-rigid warp field for real-time non-rigid reconstruction. The following work \cite{innmann2016volume, slavcheva2017killingfusion, guo2017real, yu2017BodyFusion, Slavcheva_2018_CVPR, chao2018ArticulatedFusion, yu2018doublefusion, Zheng2018HybridFusion, su2020robustfusion} incorporated different cues for more robust and accurate reconstruction. SimulCap \cite{yu2019simulcap} combined cloth simulation into the fusion pipeline but the quality of invisible details suffered from a simple cloth simulator. LiveCap \cite{habermann2019livecap} and DeepCap \cite{habermann2020deepcap} respectively solved or regressed the skeleton and non-rigid motion of a person-specific template from a monocular RGB video. TexMesh \cite{zhi2020texmesh} deformed SMPL \cite{loper2015smpl} to generate a parametric coarse mesh and generate high-quality but static texture from a single-view RGBD video.  However, because of the requirement of deforming a reference model, all these methods cannot handle topological change and reconstruct extremely non-rigid deformations.

\subsection{Tracking-free Human Inference}
Recently, more and more works focused on single RGB(D) image reconstruction because of the rise of deep learning. \cite{omran2018neural, kanazawa2018end, kolotouros2019convolutional, liang2019shape, xu2019denserac} regressed the pose and shape parameters of a human parametric model (e.g., SMPL \cite{loper2015smpl}) from a single image. Moreover, to address the challenge of general clothed human reconstruction, recent work tackled this problem by multi-view silhouettes \cite{natsume2019siclope},  depth maps \cite{gabeur2019moulding, smith2019facsimile}, template deformation \cite{zhu2019detailed, alldieck2019tex2shape}, volumetric reconstruction \cite{varol2018bodynet, zheng2019deephuman} and implicit function \cite{saito2019pifu}. DeepHuman \cite{zheng2019deephuman} conditioned single image reconstruction on the parametric SMPL model \cite{loper2015smpl} to address the problem of challenging poses. PaMIR \cite{zheng2020pamir} combined implicit function \cite{saito2019pifu} with convoluted SMPL feature for more robust and accurate inference. PIFuHD \cite{saito2020pifuhd} extended PIFu \cite{saito2019pifu} to a coarse-to-fine framework and demonstrated detailed geometric results learned from high-resolution single-view RGB images. Li \textit{et al.} \cite{li2020monocular} accelerated PIFu \cite{saito2019pifu} to achieve monocular real-time human performance capture. But all these methods focus on single frame reconstruction, but ignore temporal continuity and lack details in the invisible region.
\section{Overview}
\subsection{Preliminaries}
We firstly introduce the parametric body model \cite{loper2015smpl} and the occupancy inference network adopted in this paper.

\noindent\textbf{Parametric Body Model} The parametric body model SMPL \cite{loper2015smpl} is a function that maps the pose parameters $\theta\in\mathbb{R}^{75}$ and shape parameters $\beta\in\mathbb{R}^{10}$ to a human mesh with $N=6890$ vertices:
\begin{equation}
\begin{split}
    \label{eq: smpl formulation}
    T(\beta,\theta)=\Bar{\mathbf{T}}+B_s(\beta)+B_p(\theta),\\
    M(\beta,\theta)=W(T(\beta,\theta),J(\beta),\theta,\mathbf{W}),
\end{split}
\end{equation}
where $W(\cdot)$ is a skinning function that takes T-pose model $T(\theta,\beta)$, pose parameters $\theta$, joint positions $J(\beta)$ and skinning weights $\mathbf{W}$ as input, and returns the posed model $M(\beta,\theta)$, and $T(\theta,\beta)$ is an individual model in T-pose with shape and pose based offsets ($B_s(\beta)$ and $B_p(\theta)$).

\noindent\textbf{Occupancy Inference Network} The occupancy value $\varphi(\mathbf{x})$ of a 3D point $\mathbf{x}$ is an occupancy probability of the point inside the 3D object, and the mapping function $\varphi:\mathbb{R}^3\rightarrow \mathbb{R}$ is an implicit function.  Recently, Saito \textit{et al.} \cite{saito2019pifu} conditioned the implicit function on an image-encoded feature and proposed the pixel-aligned implicit function (PIFu):
\begin{equation}
\label{eq: pifu}
        \varphi(\mathbf{x};I)=f(G_I(\mathbf{x}_{\text{2D}}),\mathbf{x}_z),
\end{equation}
where $\mathbf{x}_{\text{2D}}=\pi(\mathbf{x})$ is the 2D projection of $\mathbf{x}$, $I$ is the conditional image, $G_I$ is a feature map of $I$ encoded by a deep image encoder, $G_I(\mathbf{x}_{\text{2D}})$ represents the sampled feature vector of $\mathbf{x}_{\text{2D}}$ on $G_I$, $\mathbf{x}_z$ is the depth value of $\mathbf{x}$, and $f(\cdot)$ is a mapping function represented by multi-layer perceptrons (MLP). Based on this pixel-level representation, PIFu can reconstruct high-fidelity details in visible regions of the object from the conditional image $I$. We improve on the original PIFu network \cite{saito2019pifu} by incorporating the feature encoded from the depth input to remove the depth ambiguity.

\begin{figure*}
    \centering
    \includegraphics[width=\textwidth]{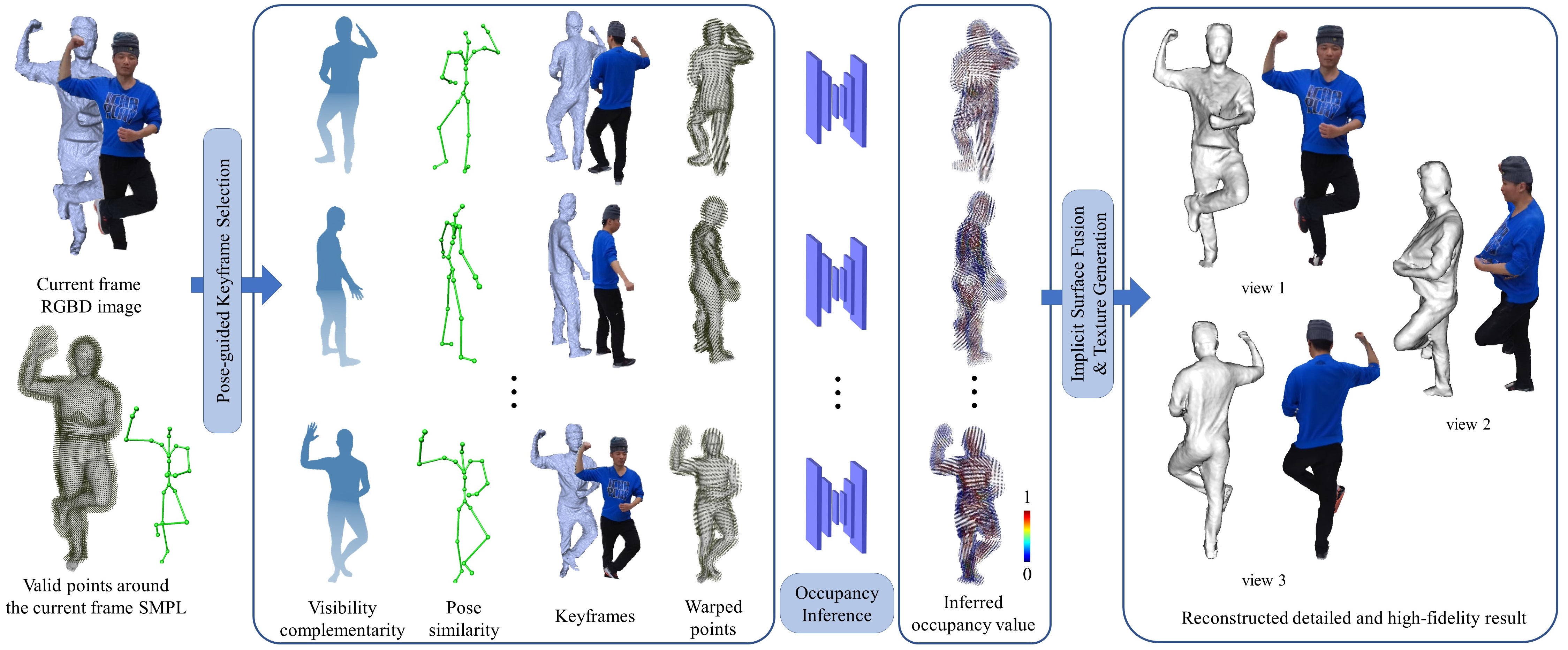}
    \caption{Reconstruction pipeline. Firstly we perform the pose-guided keyframe selection scheme to select appropriate keyframes by the visibility complementarity and pose similarity. Valid points around the current SMPL are then deformed to each keyframe by SMPL motion, and fed into a neural network with the corresponding RGBD image. The neural network infers occupancy values of each keyframe, and then we integrate all the inferred values to generate a complete model with high-fidelity and dynamic details. Finally, a high-resolution texture map is generated by projecting the reconstructed model to each keyframe RGB image and inpainted by a neural network.}
    \label{fig:pipeline}
    \vspace{-0.4cm}
\end{figure*}

\subsection{Main Pipeline}
\label{sec:pipeline}
Our goal is to reconstruct temporally continuous human models with high-quality dynamic geometric details and texture from a single-view RGBD video. At first, to construct the body motion among frames, we track the SMPL model \cite{loper2015smpl} using the whole depth sequence. For the current frame, we allocate a volume which contains the current SMPL model. We suppose that the true body surface in the current frame is near the current SMPL model, so we select valid voxels (points) around SMPL without processing redundant invalid points. Our main idea is to warp these valid points to each keyframe and fetch the corresponding occupancy values, and finally fuse a complete model with high-fidelity details. Then our system performs the following 3 steps sequentially as shown in Fig.~\ref{fig:pipeline}.
\begin{enumerate}
	\setlength{\itemsep}{0pt}
	\setlength{\parsep}{0pt}
	\setlength{\parskip}{0pt}
\vspace{-0.2cm}
	\item \textbf{Pose-guided Keyframe Selection} (Sec.~\ref{sec:keyframe selection}): To enable dynamic and high-fidelity reconstruction in the invisible region, we propose a pose-guided keyframe selection scheme that considers both visibility complementarity and pose similarity. We calculate the two metrics relative to the current frame using the tracked SMPL models, then select keyframes which not only contain the invisible regions of the current frame but are also as similar as the current SMPL pose. The pose-guided keyframe selection is further formulated as a dynamic programming to guarantee high-fidelity, dynamic and temporally continuous details in both visible and invisible regions.
	\item \textbf{Implicit Surface Fusion} (Sec.~\ref{sec:implicit surface fusion}): After the keyframe selection, the selected valid points are warped to each keyframe by SMPL motion, and then fed into a neural network to infer occupancy values which indicate the surface location contributed by this keyframe. However, the inferred values may be inaccurate. We therefore design an adaptive blending weight as the confidence and integrate occupancy values of each keyframe into the current frame to preserve high-fidelity surface details in both visible and invisible regions and guarantee smooth transition on the fusion boundaries. Moreover, if the collision occurs among different body parts, we perform the collision handling to eliminate the collision artifacts while maintaining the adjacent geometric details.
	\item \textbf{Texture Generation} (Sec~\ref{sec:texture generation}): Finally, a high-resolution texture map is generated from all the keyframes and inpainted by a neural network.
\end{enumerate}
\section{Method}

\subsection{Initialization}

Given a single-view depth stream $\{D_1,D_2,...,D_n\}$, firstly we solve the pose and shape parameters of SMPL \cite{loper2015smpl} to track each frame following the skeleton tracking of DoubleFusion \cite{yu2018doublefusion}. After that, we can obtain a SMPL sequence with pose parameters $\{\theta_1, \theta_2, ..., \theta_n\}$ corresponding to the depth stream.
For the current frame, we allocate a 3D volume which contains the SMPL model. Then valid voxels (points) are selected by the distance of each voxel to the current SMPL being less than a threshold (8cm in our experiments). What we need to do next is to solve the occupancy values of these valid points.

\begin{figure*}[t]
    \centering
    \includegraphics[width=\linewidth]{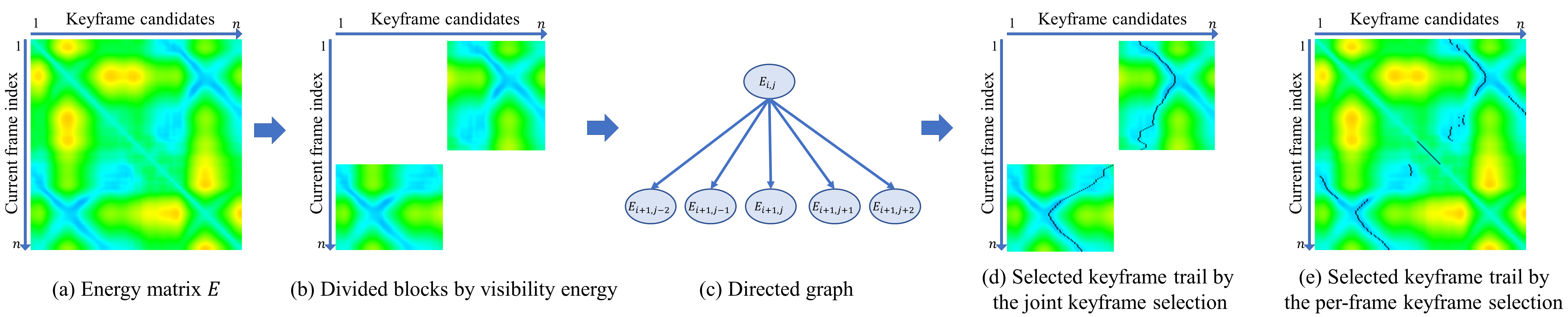}
    \caption{Illustration of the two solutions to the keyframe selection. (a)(b)(c)(d) We formulate the keyframe selection as a dynamic programming (DP) problem to select temporally continuous keyframes, (e) the keyframe trail by the per-frame keyframe selection.}
    \label{fig: pose-dependent keyframe selection}
    \vspace{-0.4cm}
\end{figure*}

\subsection{Pose-guided Keyframe Selection}
\label{sec:keyframe selection}
Consider the $i$-th frame as the current frame, the proposed pose-guided keyframe selection chooses appropriate frames by both pose similarity and visibility complementarity from other frames $F=\{1,2,...,i-1,i+1,...,n\}$. The parametric SMPL model \cite{loper2015smpl} across the whole sequence contributes to quantify visibility complementarity and pose similarity. More importantly, the keyframe selection should guarantee the temporal continuity of the selected keyframes between adjacent frames. In the keyframe selection, our goal is to select $K$ ($K=4$ in our experiments) keyframes from $F$ by our proposed pose and visibility metrics, and in each iteration, we select one keyframe.

\noindent\textbf{Pose Similarity} Based on the parametric SMPL model, we can formulate the pose similarity energy between the $j$-th frame and the current frame conveniently as:
\begin{equation}
    \label{eq: pose similarity energy}
    E_{\text{pose}}(i,j) = \sum_{k\in\mathcal{J}}w_k|\theta_i^k-\theta_j^k|^2,
\end{equation}
where $\mathcal{J}$ is the joint index set except the global rotation and translation, and $w_k$ is the influence weight of the $k$-th joint to the keyframe selection. The pose similarity constrains that the body pose in the selected keyframe is similar to that in the current frame.

\noindent\textbf{Visibility Complementarity} With the topology-consistent SMPL model across the whole sequence, we can set a visible flag for each vertex of SMPL in the $j$-th frame:
\begin{equation}
f_j^k = \left\{\begin{matrix}
1 ,& \text{$\mathbf{v}_k$ is visible in the $j$-th frame}\\ 
0 ,& \text{$\mathbf{v}_k$ is not visible in the $j$-th frame} 
\end{matrix}\right.,
\end{equation}
where $\mathbf{v}_k$ is the $k$-th vertex of SMPL. So we can define a visible vector $\mathbf{f}_j=[f_j^1,f_j^2,\cdots,f_j^N]^\top$ which encodes the visible region of human body for each frame. The selected keyframe set of the $i$-th frame in previous iterations is denoted as $\mathcal{K}_i$, and the visibility $\mathbf{F}_{\mathcal{K}_i}$ over $\mathcal{K}_i$ is defined as:
\begin{equation}
    \mathbf{F}_{\mathcal{K}_i}=\lor_{k\in \mathcal{K}_i} \mathbf{f}_k,
\end{equation}
where $\lor$ is the element-wise logical OR operation. Before the first iteration, we initialize $\mathcal{K}_i=\{i\}$. In each iteration, the visibility complementarity energy is defined as:
\begin{equation}
\label{eq: visibility energy}
    E_{\text{visibility}}(\mathcal{K}_i, j) = \frac{\left\| \neg\mathbf{F}_{\mathcal{K}_i} \land \mathbf{f}_j \right\|_0}{\|\neg \mathbf{F}_{\mathcal{K}_i}\|_0},
\end{equation}
where $\neg$ and $\land$ are element-wise logical NOT and AND operations respectively, and Eq.~\ref{eq: visibility energy} represents the proportion of ``new'' visible vertices that are visible in the $j$-th frame but not in $\mathcal{K}_i$ to all the invisible vertices in $\mathcal{K}_i$.

\noindent\textbf{Joint Keyframe Selection} In each iteration, we construct an energy matrix $E\in\mathbb{R}^{n\times n}$ ($n$ is the frame number), and the $(i,j)$-th element of $E$ is defined as
\begin{equation}
    E_{i,j}=E_{\text{pose}}(i,j)-\lambda_{\text{visibility}} E_{\text{visibility}}(\mathcal{K}_i,j),
\end{equation}
where $\lambda_{\text{visibility}}$ is a term weight. The energy matrix $E$ encodes the visibility complementarity and pose similarity between the current frame and each keyframe candidate as shown in Fig.~\ref{fig: pose-dependent keyframe selection}(a). We consider the row and column indices of $E$ as the current frame index and keyframe candidates, respectively. We define the selected keyframe trail $\mathcal{T}$ in each iteration as $\mathcal{T}=\{t_1,t_2,...,t_n\}$,
where $t_i$ is the selected keyframe of the $i$-th frame. A potential solution of the keyframe selection is to select the minimal element of each row, i.e., $t_i=\mathop{\arg\min}_{j}E_{i,j}$. However, the drawback of the per-frame selection is the temporal discontinuity of selected keyframes between the adjacent frames (Fig.~\ref{fig: pose-dependent keyframe selection}(e)).

To guarantee the temporal continuity of details in invisible regions, we jointly select keyframes for the whole sequence, and formulate this procedure as a dynamic programming (DP) problem as illustrated in Fig.~\ref{fig: pose-dependent keyframe selection}(b, c, d). To avoid the continuous keyframe trial to cross the diagonal of $E$, we firstly utilize visibility energy to divide several blocks and select keyframes within each block independently as shown in Fig.~\ref{fig: pose-dependent keyframe selection}(b). For two adjacent blocks, we maintain a FIFO keyframe queue for smooth transition. Within each block, we constrain that the selected frames between two adjacent frames should be temporally continuous with each other, i.e., $|t_{i+1}-t_{i}|\leq \tau$,
where $\tau$ is a half window size, and $\tau=2$ in Fig.~\ref{fig: pose-dependent keyframe selection}(c). Then we connect $E_{i,j}$ with $\{E_{i+1,j-\tau},...,E_{i+1,j+\tau}\}$ respectively to construct a directed graph (Fig.~\ref{fig: pose-dependent keyframe selection}(c)), and our goal is to find a trail from the first row to the last row with the minimal energy sum for a global optimal solution, which is naturally a dynamic programming problem of minimum cost path. Based on this formulation, we can obtain a temporally continuous keyframe trail as illustrated in Fig.~\ref{fig: pose-dependent keyframe selection}(d).

\begin{figure*}[t]
    \centering
    \includegraphics[width=\linewidth]{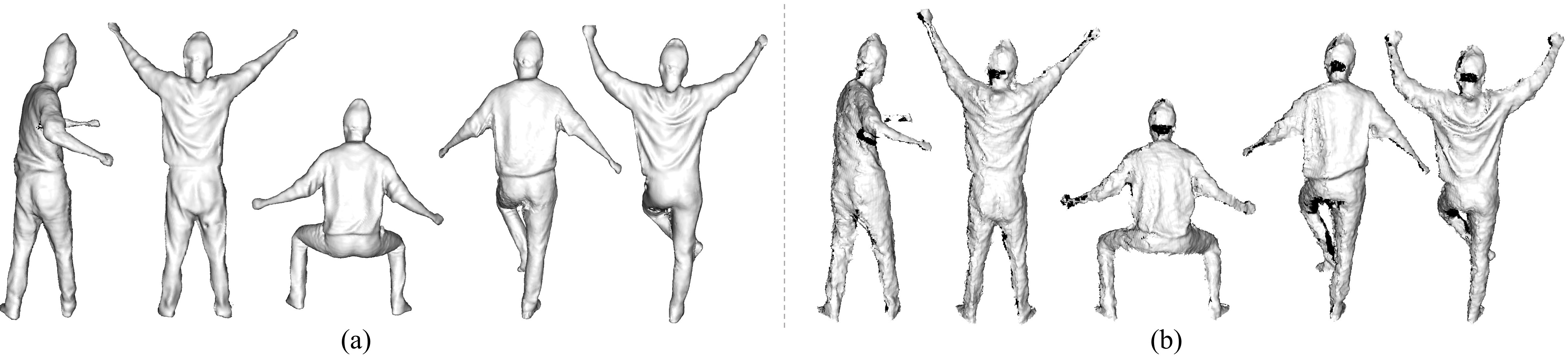}
    \caption{Comparison of reconstructed invisible details (a) and the ground truth (b) captured by multiple calibrated Kinects. These dynamic invisible details are similar to the physical ones due to the pose guidance.}
    \label{fig:pose-dependent details}
    \vspace{-0.4cm}
\end{figure*}

After each iteration, we update the selected keyframe set of each frame and the visibility over it:
\begin{equation}
    \mathcal{K}_i\leftarrow\mathcal{K}_i\cup \{t_i\},\,
   \mathbf{F}_{\mathcal{K}_i}\leftarrow\mathbf{F}_{\mathcal{K}_i} \lor \mathbf{f}_{t_i},\, i=1,...,n.
\end{equation}
In the next iteration, we construct the energy matrix $E$ using the updated $\{\mathcal{K}_i\}_{i=1}^n$ and $\{\mathbf{F}_{\mathcal{K}_i}\}_{i=1}^n$ to search another continuous keyframe trail to cover other invisible regions.

Based on the proposed pose-guided keyframe selection, our method can reconstruct dynamic and high-fidelity details in the invisible regions as shown in Fig.~\ref{fig:pose-dependent details}.

\subsection{Implicit Surface Fusion}
\label{sec:implicit surface fusion}
\noindent\textbf{Occupancy Inference} After the keyframe selection, we have a keyframe set $\mathcal{K}_i$ for the current $i$-th frame. For each keyframe $k\in\mathcal{K}_i$, we firstly deform the valid points from the current frame to the $k$-th frame by the SMPL motion, then feed them to the occupancy inference network with the corresponding RGBD image, and finally obtain the occupancy values contributed by this keyframe.

\noindent\textbf{Adaptive Blending Weight} The inferred occupancy values may be inaccurate in invisible regions especially for self-occluded input or challenging poses. So directly averaging inferred occupancy values provided by all the keyframes just like in DynamicFusion \cite{newcombe2015dynamicfusion} is improper. Our observation is that thanks to the depth information, PIFu can provide quite precise inference near the depth point clouds and in the visible region. We therefore design an adaptive blending weight according to visibility and the distance between each valid point and the depth point clouds, and the weight is formulated as:
\begin{equation}
\label{eq: blending weight}
\begin{split}
w(\mathbf{x};D_k)=&
\left\{\begin{matrix}
1 & ,p(\mathbf{x};D_k) < \tau\\ 
e^{-\sigma(p(\mathbf{x};D_k) - \tau)} & ,p(\mathbf{x};D_k) \geq \tau
\end{matrix}\right.,\\
&p(\mathbf{x};D_k)=\mathbf{x}_z-D_k(\pi(\mathbf{x})),
\end{split}
\end{equation}
where $p(\mathbf{x};D_k)$ is the projected signed distance function (PSDF) \cite{curless1997new} that takes a 3D point $\mathbf{x}$ as input and returns the difference between the depth value of $\mathbf{x}$ and the sampled depth value at the projected location $\pi(\mathbf{x})$ on the depth image $D_k$, $\sigma$ is a factor to control the descending speed of the blending weight along the projection direction, and $\tau$ is a threshold to define high-confident regions. Then we integrate the inferred occupancy values from each keyframe using Eq.~\ref{eq: blending weight}, and finally extract a model with high-fidelity geometric details using Marching Cubes \cite{lorensen1987marching}.

\begin{figure}[t]
    \centering
    \includegraphics[width=\linewidth]{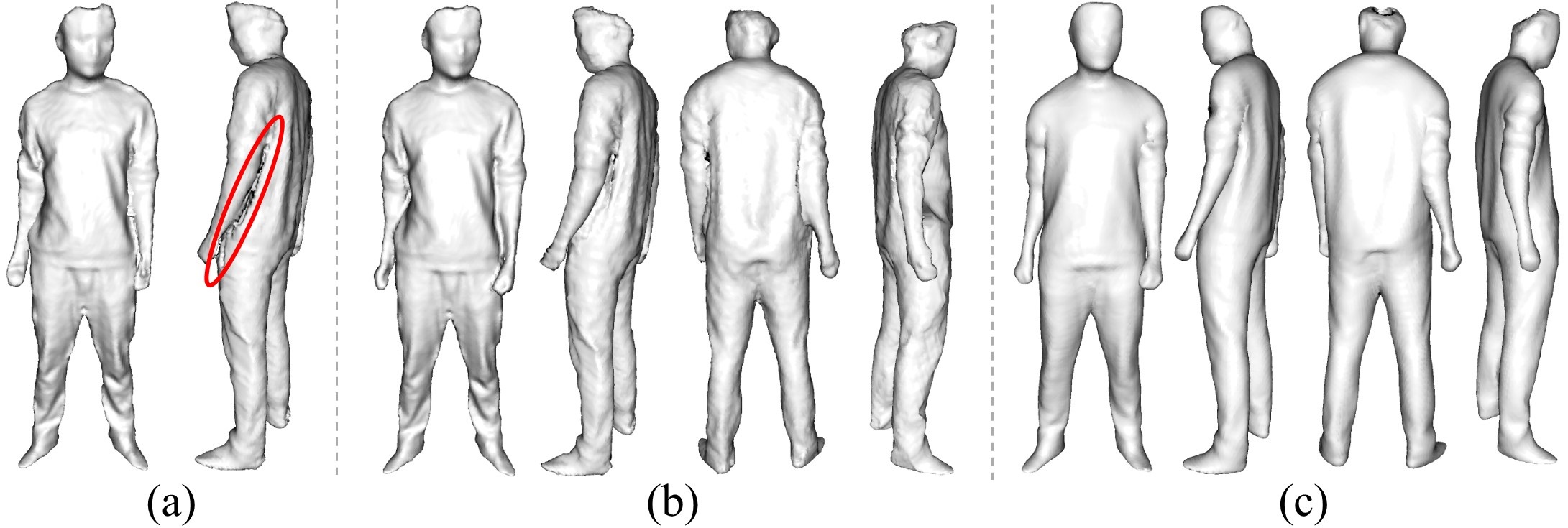}
    \caption{Illustration of the collision handling. (a) Reconstructed model without collision handling, (b) reconstructed model by fusing an adjacent no-collision model, (c) reconstructed model by fusing a continuously maintained template.}
    \label{fig:eval collision}
    \vspace{-0.4cm}
\end{figure}

\noindent\textbf{Collision Handling} Though the pipeline has been carefully designed to achieve dynamic and high-fidelity reconstruction, it still suffers from the collision problem. The self-collision in the live frame has been studied in \cite{dou2016fusion4d, guo2017real}, however, the collision in the reference frame (i.e., the current frame in our method) is urgent to be resolved. For example, consider that the left arm of the performer collides with the torso as shown in Fig.~\ref{fig:eval collision}(a), it is confusing to decide which body part to drive these collided points around this region. Due to the tracking error and the difference between SMPL and the real clothed human body, these points may be warped to incorrect positions and fetch wrong occupancy values, so that a crack occurs in the collided region as shown in the red ellipse of Fig.~\ref{fig:eval collision}(a). A potential solution is to follow some tracking-based methods \cite{yu2018doublefusion, Zheng2018HybridFusion, su2020robustfusion} in which they maintain a no-collision model under A-pose or T-pose, and deform the no-collision model to another frame. However, the continuously maintained model is being oversmoothed and loses the adjacent details due to the continuous fusion.
Our observation is that since the body motion is continuous, a no-collision reconstructed model $\mathbf{M}^\prime$ exists near the collided current frame. So we can deform this adjacent and no-collision model to the current frame, and voxelize it into an occupancy field $\mathbf{O}^\prime$, and finally integrate $\mathbf{O}^\prime$ into the implicit surface fusion. As shown in Fig.~\ref{fig:eval collision}(b)(c), our collision handling scheme not only eliminates the collision artifacts but also maintains the geometric details in other regions, while fusing a continuous model loses the adjacent details.

\begin{figure*}[t]
    \centering
    \includegraphics[width=\linewidth]{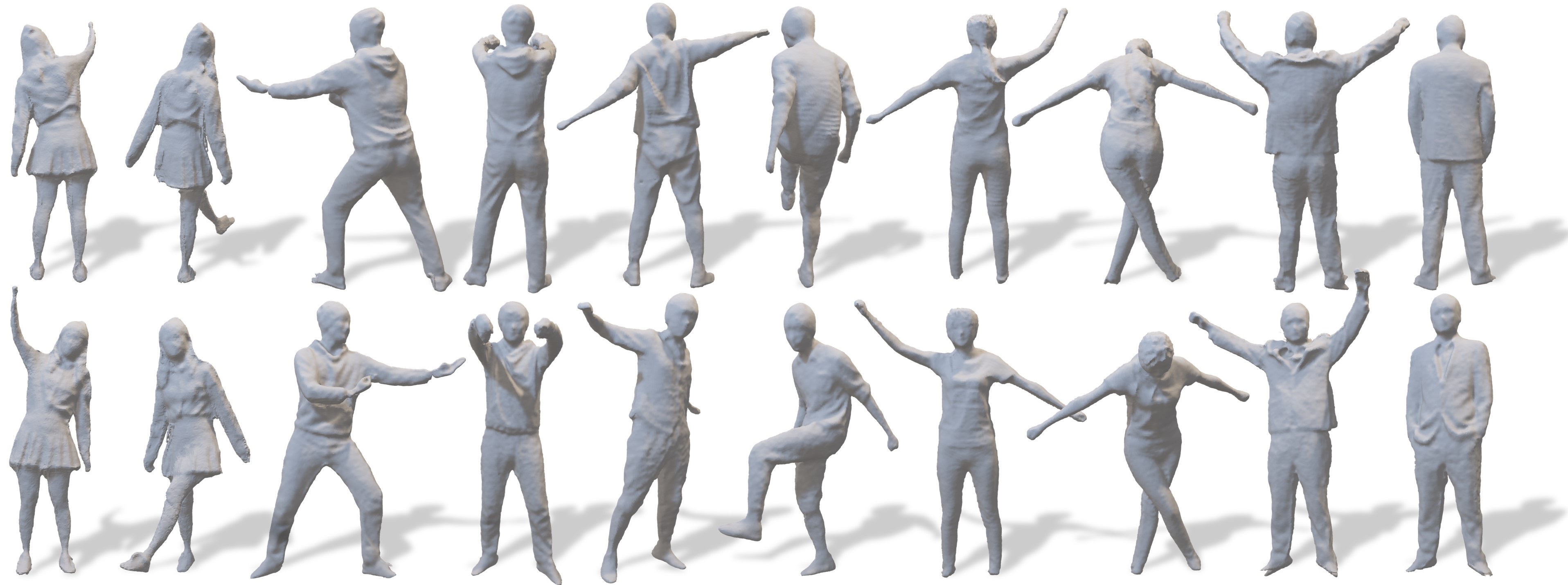}
    \caption{Results with dynamic and high-fidelity details reconstructed by our method. The bottom row is the input view, and the top row is another rendering view. Dynamic and high-fidelity details are reconstructed in both visible and invisible regions.}
    \label{fig:results}
    \vspace{-0.4cm}
\end{figure*}

\begin{figure*}[t]
    \centering
    \includegraphics[width=\linewidth]{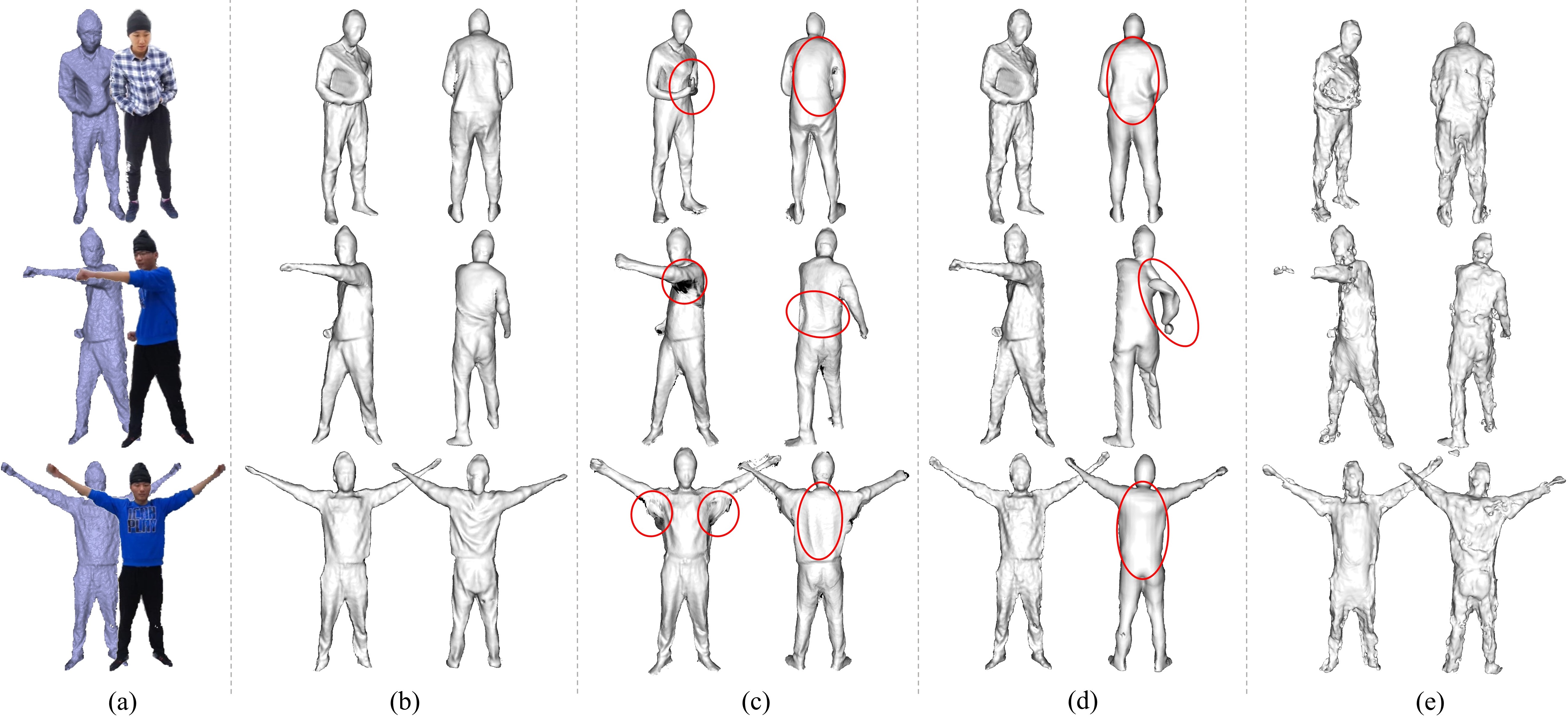}
    \caption{Qualitative comparison against other state-of-the-art methods. (a) RGBD images in the current frame, and results by our method (b), DoubleFusion \cite{yu2018doublefusion} (c), RGBD-PIFu \cite{saito2019pifu} (d) and IP-Net \cite{bhatnagar2020ipnet} (e).}
    \label{fig:qualitative comparison}
    \vspace{-0.5cm}
\end{figure*}

\subsection{Texture Generation}
\label{sec:texture generation}
Given the geometric model $\mathbf{M}_i$ in the current frame, we utilize a per-face tile \cite{soucy1996texture} to represent the texture on each face of $\mathbf{M}_i$ for high-resolution texture and fast UV unwrapping. Then we deform $\mathbf{M}_i$ to each keyframe and project it to the RGB image to fetch RGB values and finally blend them together. For the invisible faces, we utlize a neural network \cite{saito2019pifu} to infer their textures.

\section{Results}
In this section, we firstly compare our method with current state-of-the-art works qualitatively and quantitatively. Then we evaluate our main contributions. Some results captured by our system are demonstrated in Fig.~\ref{fig:results}. Please refer to the supplemental material for the implementation details.

\subsection{Comparison}
\begin{figure}[t]
    \centering
    \includegraphics[width=\linewidth]{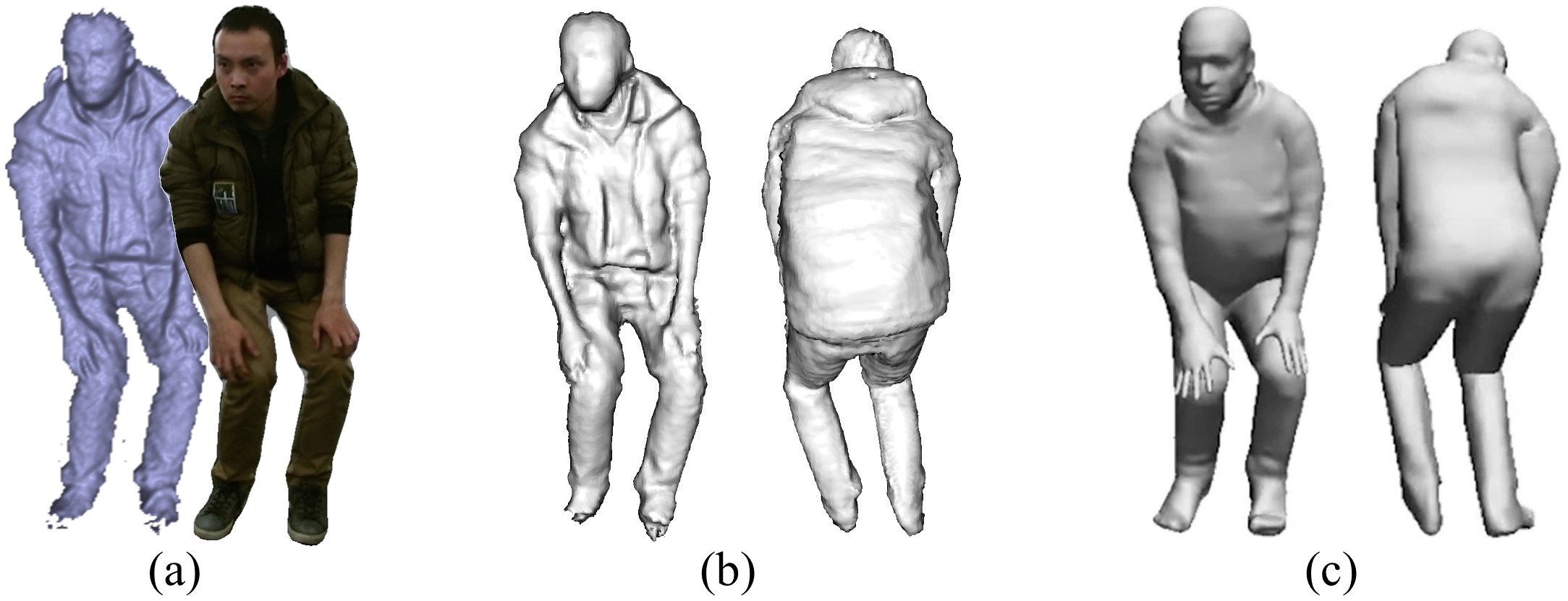}
    \caption{Comparison against TexMesh \cite{zhi2020texmesh}. (a) RGBD image in the current frame, and results by our method (b) and TexMesh (c).}
    \label{fig:cmp with texmesh}
    \vspace{-0.2cm}
\end{figure}

\noindent\textbf{Qualitative Comparison} We compare the geometric reconstruction of our method with some representative and state-of-the-art tracking-based \cite{yu2018doublefusion} and tracking-free \cite{saito2019pifu} works as well as 3D human completion method \cite{bhatnagar2020ipnet} qualitatively using our captured data by a Kinect Azure in Fig.~\ref{fig:qualitative comparison}.
And our method outperforms these methods on topological changes (top row of Fig.~\ref{fig:qualitative comparison}(c)), natural deformations (middle and bottom rows of Fig.~\ref{fig:qualitative comparison}(c)), dynamic pose-guided details (Fig.~\ref{fig:qualitative comparison}(c, d)) and invisible details (Fig.~\ref{fig:qualitative comparison}(d)). For a fair comparison, we retrain PIFu \cite{saito2019pifu} with depth inputs and denote it as RGBD-PIFu, and the input of IP-Net \cite{bhatnagar2020ipnet} is a roughly complete depth point cloud merged from keyframes. However, due to the skeleton-level deformation error and depth noise, IP-Net fails to recover a complete detailed human model. We also compare our method with TexMesh \cite{zhi2020texmesh} using their data in Fig.~\ref{fig:cmp with texmesh}, and our method can reconstruct much more detailed geometry.

\begin{figure}[t]
    \centering
    \includegraphics[width=\linewidth]{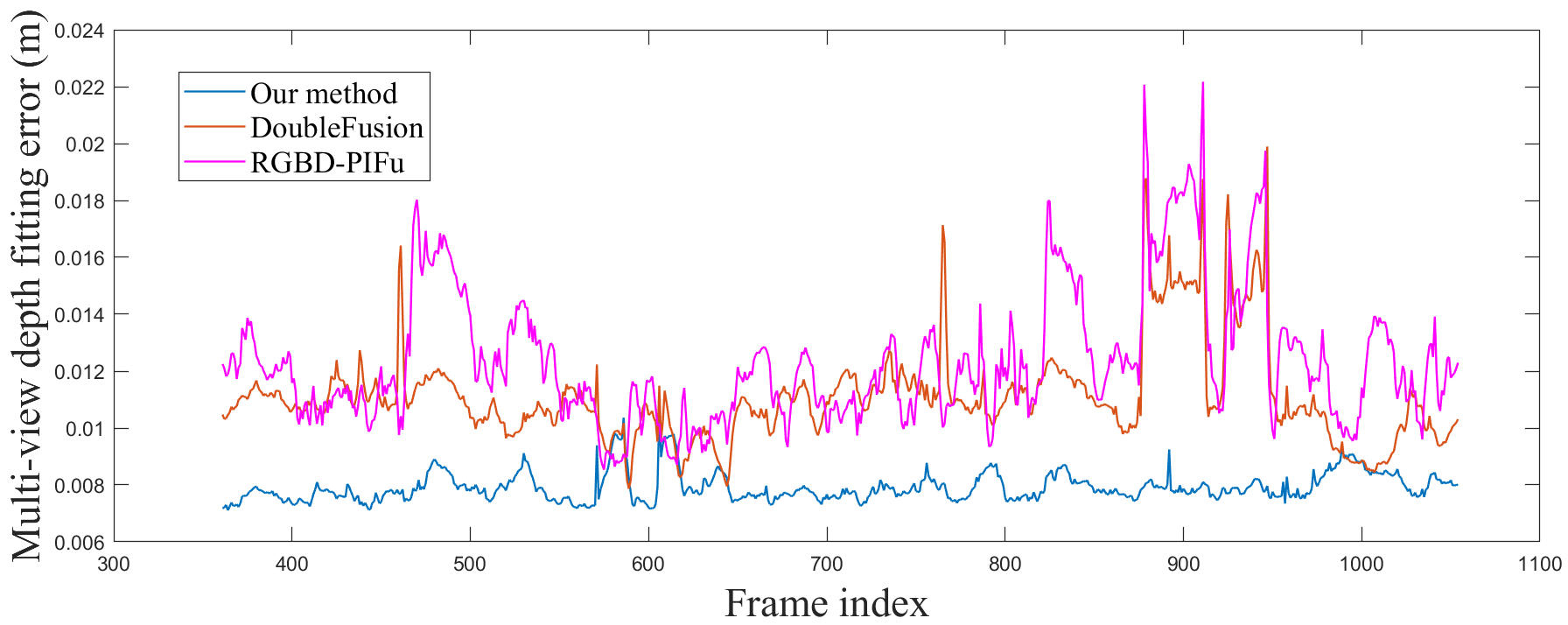}
    \caption{Quantitative comparison on the multi-view depth fitting error of our method, DoubleFusion \cite{yu2018doublefusion} and RGBD-PIFu \cite{saito2019pifu}.}
    \label{fig:quantitative comparison}
    \vspace{-0.2cm}
\end{figure}

\noindent\textbf{Quantitative Comparison}
We compare our method with DoubleFusion \cite{yu2018doublefusion} and RGBD-PIFu \cite{saito2019pifu} on the multi-view depth fitting error quantitatively. We utilize 4 calibrated Kinects to capture multi-view point clouds as the target, and evaluate the mean fitting error of each frame as shown in Fig.~\ref{fig:quantitative comparison}. It shows that our method reconstructs much more accurate results than DoubleFusion and RGBD-PIFu because in our method the invisible regions are similar to the physical ones thanks to the pose guidance and the visible regions are exactly same as the current observation.

\subsection{Evaluation}
\begin{figure}[t]
    \centering
    \includegraphics[width=\linewidth]{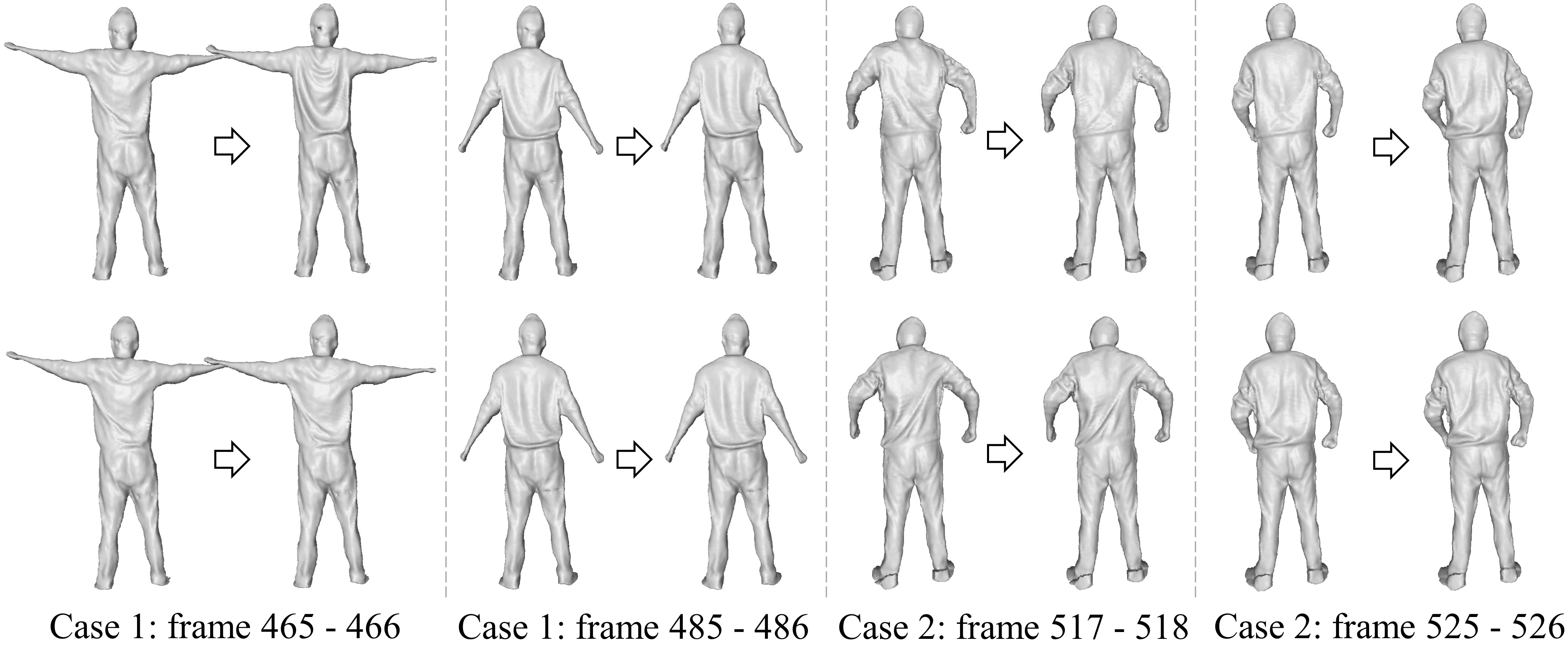}
    \caption{Comparison against per-frame keyframe selection. Invisible details of several adjacent frames by joint keyframe selection (bottom row) and per-frame selection (top row), respectively.
    }
    \label{fig: against per-frame kf selection}
    \vspace{-0.4cm}
\end{figure}

\begin{figure}[t]
    \centering
    \includegraphics[width=\linewidth]{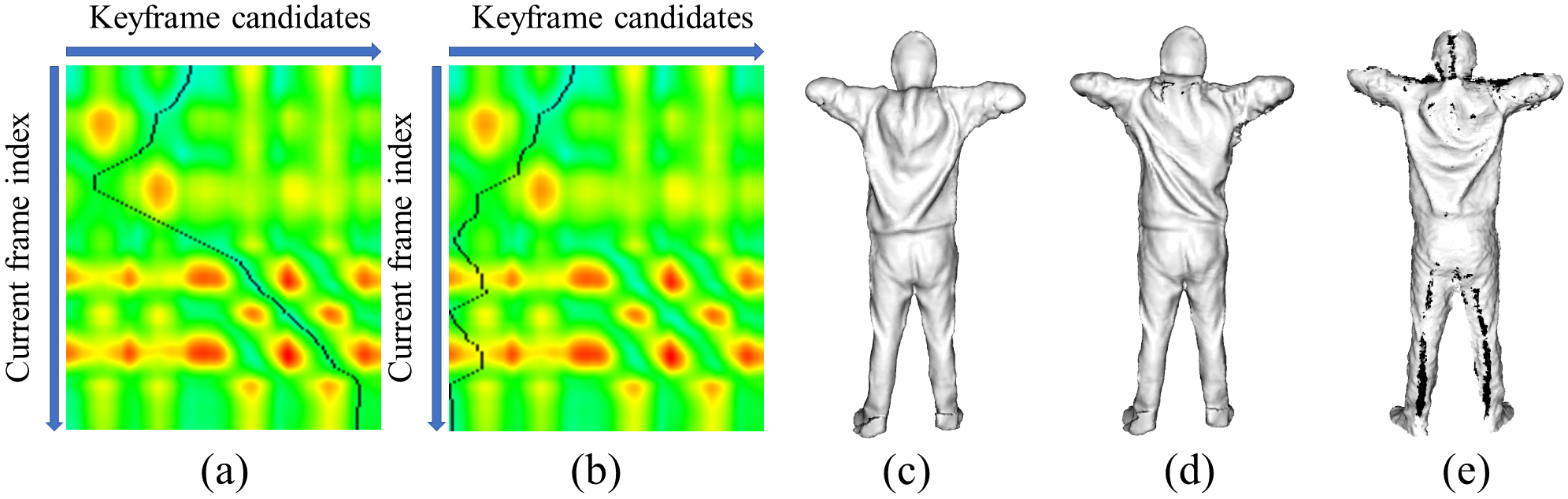}
    \caption{Comparison against the greedy algorithm. (a)(b) Keyframe trails obtained by DP and greedy algorithm, respectively, (c)(d) invisible details reconstructed by DP and greedy algorithm, respectively, (e) ground truth of the invisible regions.}
    \label{fig: against greedy algorithm}
    \vspace{-0.2cm}
\end{figure}

\noindent\textbf{Joint Keyframe Selection}\\
-- \textbf{Comparison against Per-frame Keyframe Selection} We demonstrate invisible details of several adjacent frames reconstructed with the joint keyframe selection and per-frame keyframe selection in Fig.~\ref{fig: against per-frame kf selection}, respectively. 
It shows the superiority of the joint keyframe selection on the temporal coherence compared with the per-frame keyframe selection.\\
-- \textbf{Comparison against Greedy Algorithm} We compare the dynamic programming (DP) solution against the greedy algorithm\footnote{The greedy algorithm: considering the directed graph constructed in Fig.~\ref{fig: pose-dependent keyframe selection}(c), given the keyframe $t_i$ of the $i$-th frame, the keyframe $t_{i+1}$ of the next frame is selected using $t_{i+1}=\mathop{\arg\min}_{j\in\{t_i-\tau,\cdots,t_i+\tau\}}E_{i,j}$.} in Fig.~\ref{fig: against greedy algorithm}. It shows that even though the greedy algorithm can obtain a temporally continuous keyframe trail, this algorithm may fall into a local minimum and some frames select keyframes with high energies in which the human poses are not similar to the current ones, so that the reconstructed invisible details are not physically plausible.

\begin{figure}[t]
    \centering
    \includegraphics[width=\linewidth]{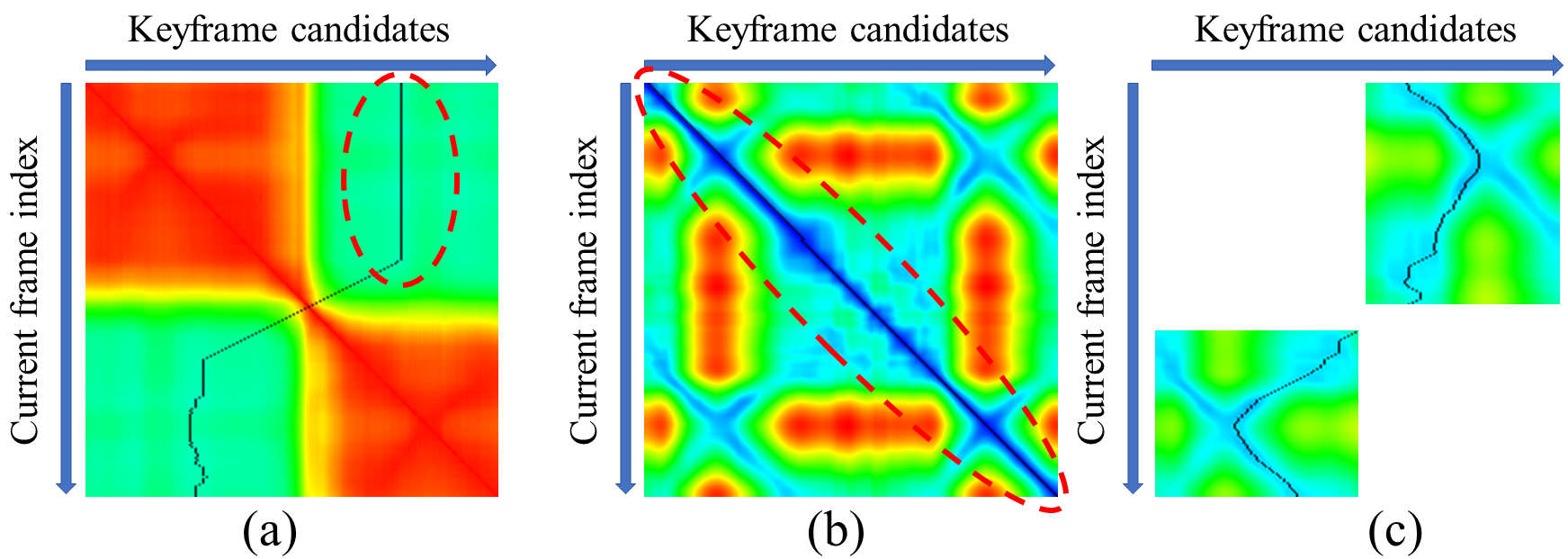}
    \caption{Qualitative ablation study of the pose-guided keyframe selection. 
    (a) The keyframe trail without the pose energy, (b) the keyframe trail without the visibility energy, (c) the keyframe trail using the total energy within each divided block.}
    \label{fig:ablation study kf selection}
    \vspace{-0.2cm}
\end{figure}

\begin{figure}[t]
    \centering
    \includegraphics[width=\linewidth]{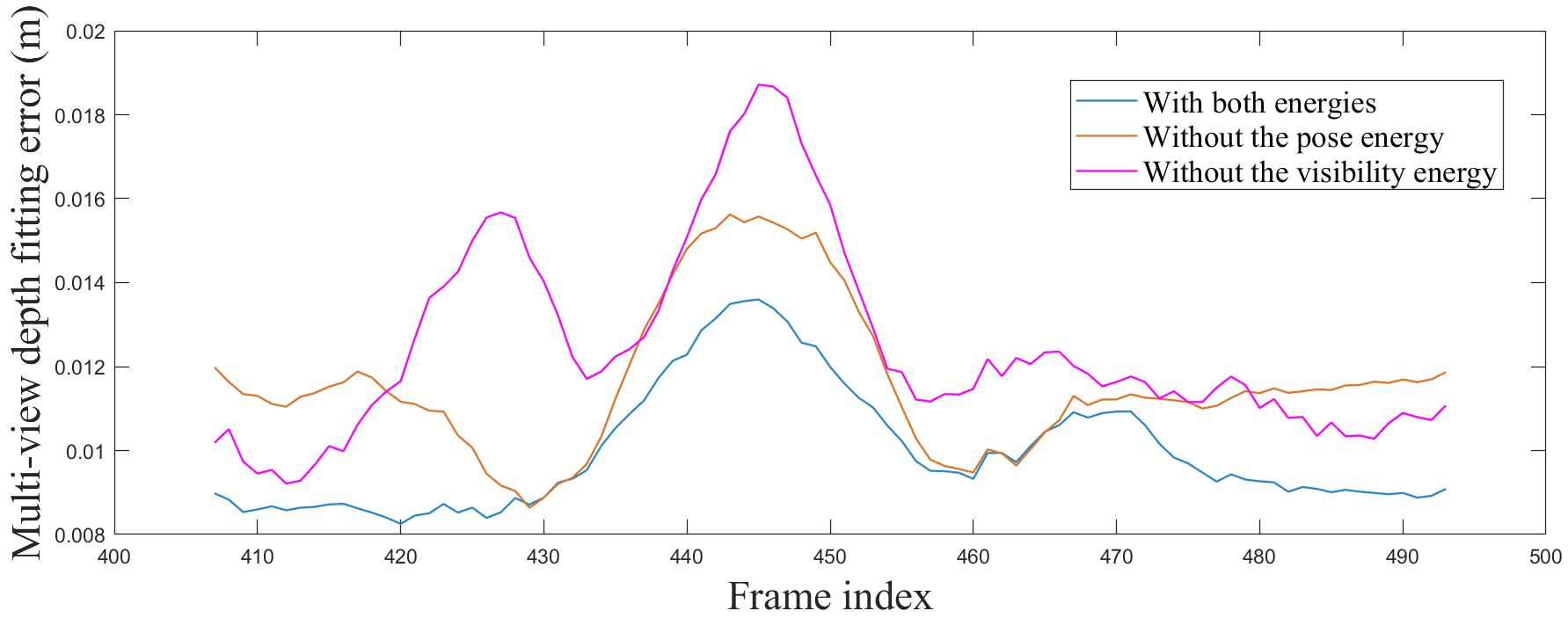}
    \caption{Quantitative ablation study of the pose-guided keyframe selection on the mean multi-view depth fitting error.}
    \label{fig:quantitative ablation study kf selection}
    \vspace{-0.4cm}
\end{figure}

\noindent\textbf{Ablation Study of Pose-guided Keyframe Selection} We eliminate the pose or visibility energy, and visualize the keyframe trail in each situation in Fig.~\ref{fig:ablation study kf selection}. \textbf{Pose Energy:} The red ellipse in Fig.~\ref{fig:ablation study kf selection}(a) demonstrates that these frames select the same keyframe, so that the reconstructed invisible details in these frames are static. \textbf{Visibility Energy:} Fig.~\ref{fig:ablation study kf selection}(b) demonstrates that without the visibility energy, the pose energy guides the selection scheme to choose the current frames, which is irrational. Fig.~\ref{fig:ablation study kf selection}(c) shows that with both energies the keyframe selection provides a temporally continuous keyframe trail to generate dynamic pose-guided details in the invisible regions. For each situation (Fig.\ref{fig:ablation study kf selection}(a, b, c)), we evaluate the multi-view depth fitting error quantitatively as shown in Fig.~\ref{fig:quantitative ablation study kf selection}, and using both energies reconstructs more accurate geometry because the pose-guided invisible details are similar to the physical ones.

\section{Discussion}
\noindent\textbf{Conclusion}
In this paper, we propose Pose-guided Selective Fusion (POSEFusion), the first method that can reconstruct high-fidelity and dynamic details of a performer even in the invisible regions from only a single RGBD camera. Based on the proposed pose-guided selective fusion framework, our method effectively combines the advantages of tracking-based methods and tracking-free inference methods while avoiding their drawbacks. As a result, our method outperforms the other state-of-the-art monocular capture methods. 

\noindent\textbf{Limitation and Future Work}
It remains difficult for our method to handle very loose cloth (e.g., long skirt) because it is challenging to track a performer wearing loose cloth only using SMPL \cite{loper2015smpl}. Replacing SMPL with a pre-scanned template (e.g., \cite{li20133d, li2020robust}) may remove this limitation. Moreover, keyframe candidates in our keyframe selection are required to be sequential to guarantee temporal coherence, as for a non-sequential database, reorganizing these candidates by shape similarity \cite{budd2013global} may resolve this problem. In addition, the reconstructed invisible details may not be exactly the same as the real ones, which we leave for future research. 

\noindent\textbf{Acknowledgement} This paper is supported by the National Key Research and Development Program of China [2018YFB2100500], the NSFC No.61827805 and No.61861166002, and China Postdoctoral Science Foundation No.2020M670340.

\clearpage
\appendix
\noindent\textbf{\large Supplementary Material}\\

In this supplementary material, we mainly introduce our implementation details, the system performance, and additional experiments.
\section{Implementation Details}
\subsection{Initialization}
Given the tracked SMPL model \cite{loper2015smpl} in the current frame, we firstly allocate a 3D volume which contains the SMPL. For each voxel, we calculate the Euclidean distance between its center with SMPL, and select valid voxels (points) near the SMPL with a threshold of 8cm.

\subsection{Pose-guided Keyframe Selection}
\begin{figure}[t]
    \centering
    \includegraphics[width=\linewidth]{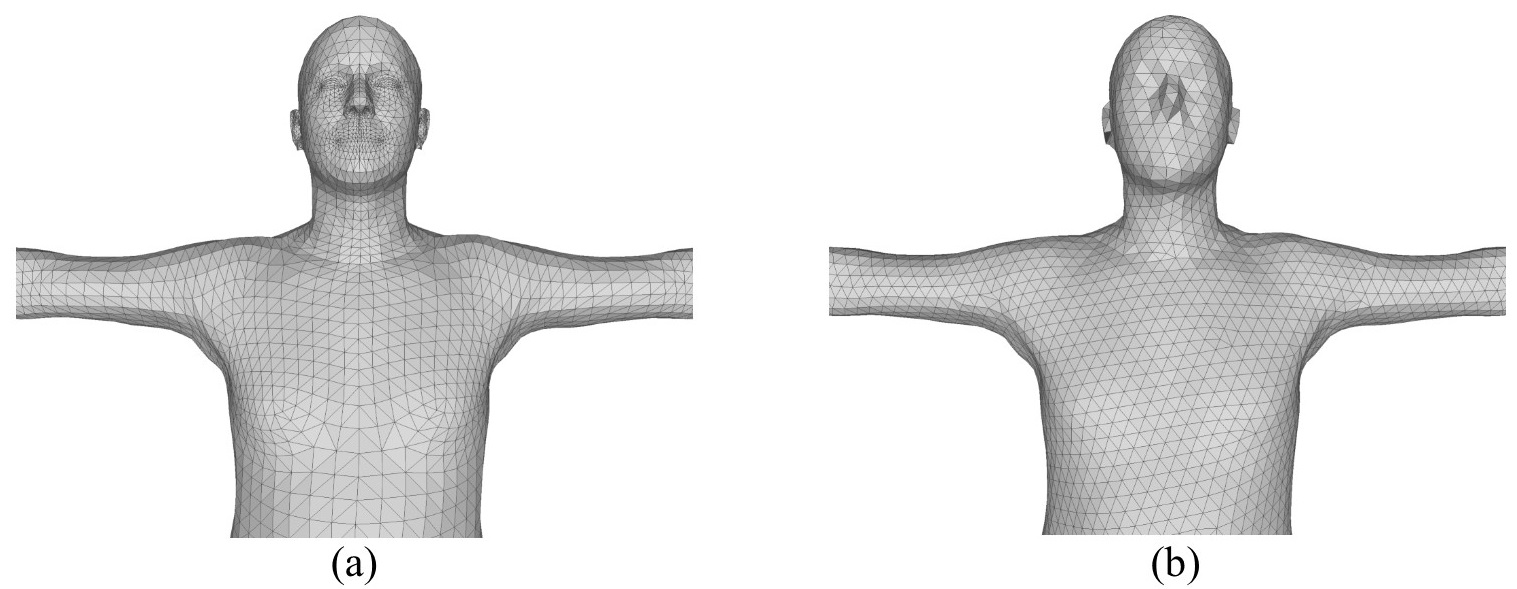}
    \caption{SMPL remeshing using \cite{jakob2015instant}. (a)(b) The wireframes of the original SMPL and the new template, respectively.}
    \label{fig:smpl remeshing}
    \vspace{-0.4cm}
\end{figure}

\noindent\textbf{SMPL Remeshing} Due to the uneven distribution of vertices of the original SMPL \cite{loper2015smpl}, we remesh the original SMPL into an isotropic triangular template using \cite{jakob2015instant} as shown in Fig.~\ref{fig:smpl remeshing}. We then transfer the parameters (e.g., blending weights) in \cite{loper2015smpl} to the new template. Using the new template with uniform triangles, the visibility energy can be calculated without the negative impact of dense vertices in the face and hand regions as shown in Fig.~\ref{fig:smpl remeshing}(a). In our experiment, the new template contains 6839 vertices and 13690 faces. For simplicity, we also call the new template SMPL in the main paper and this supplementary material.

\begin{figure}[t]
    \centering
    \includegraphics[width=\linewidth]{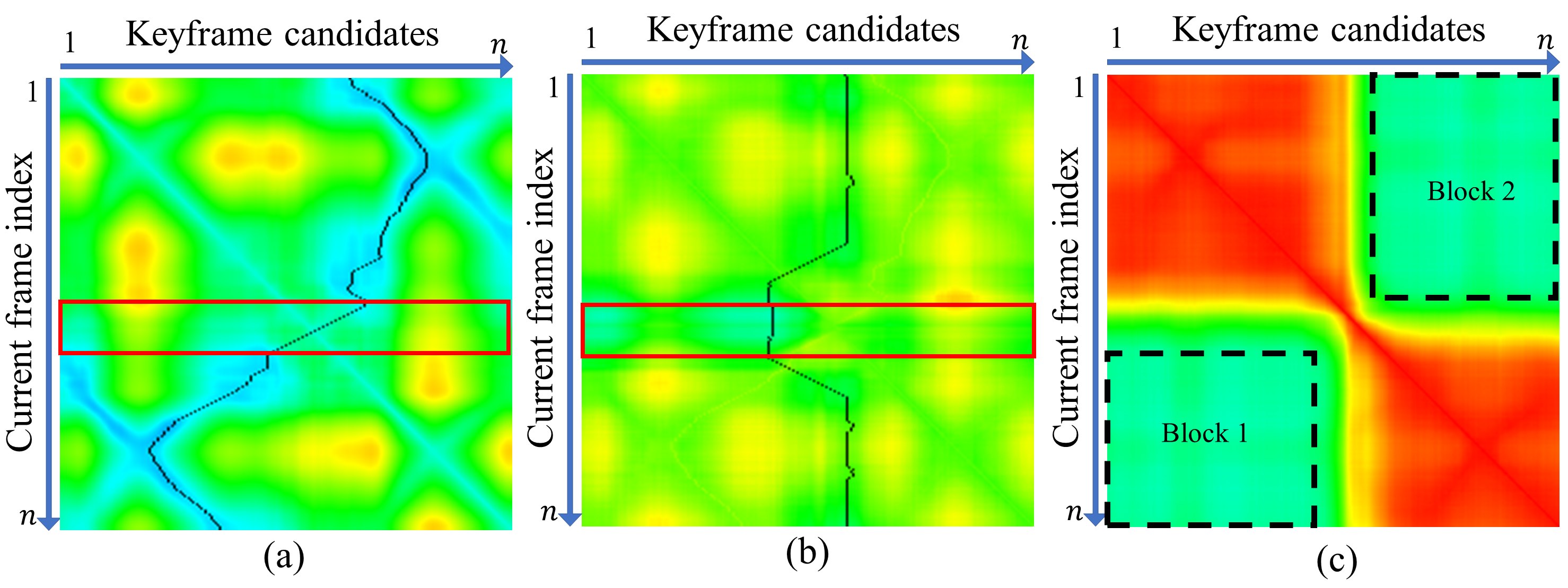}
    \caption{Illustration of the block division by the visibility energy. (a)(b) the keyframe trail on the energy matrix $E$ in the first and second iteration without block division, respectively, and in the red rectangle the selected keyframe in the first iteration is very close to the current frame; (c) the divided blocks by the visibility matrix $E_{\text{v}}$ of the first iteration.}
    \label{fig:block-wise refine}
    \vspace{-0.4cm}
\end{figure}

\noindent\textbf{Block Division by Visibility}
Though the formulated dynamic programming (DP) solution can provide a temporally continuous keyframe trail, this trail in the first iteration denoted as $\mathcal{T}^1$ may cross the diagonal region of the energy matrix $E$ (the red rectangle in Fig.~\ref{fig:block-wise refine}(a)), which indicates that in this region the selected keyframe $t_i$ is very close to the current $i$-th frame, i.e., the $i$-th and $t_i$-th frames have roughly the same visible region of the human body. In the second iteration, the DP generates another trail denoted as $\mathcal{T}^2$ as shown in Fig.~\ref{fig:block-wise refine}(b). It is obvious that for the red rectangle region $\mathcal{T}^2$ has more contribution for the visibility complementarity, however, $\mathcal{T}^2$ may be discontinuous with $\mathcal{T}^1$ on the boundary of the red rectangle. This will deteriorate the temporal continuity of the reconstructed details in invisible regions in this rectangle region.

Our observation is that the red rectangle usually occurs when the visible region changes significantly, e.g., the performer turns from facing the camera to back facing it, which can be intuitively seen from the visibility matrix $E_{\text{v}}$ of the first iteration (the $(i,j)$-th element of $E_{\text{v}}$ is $1-E_{\text{visibility}}(\mathcal{K}_i,j)$) as shown in Fig.~\ref{fig:block-wise refine}(c). We therefore firstly divide the whole sequence as several blocks using  $E_{\text{v}}$ as shown in Fig.~\ref{fig:block-wise refine}(c). If the trail crosses two individual blocks, the red rectangle will be bounded, then we maintain a fixed-size FIFO (first-in-first-out) queue $\mathcal{Q}$ for the rectangle region and push the selected keyframes of previous frames into $\mathcal{Q}$. $\mathcal{Q}$ is considered as the keyframe set of the red rectangle region to guarantee the temporally smooth transition of the reconstructed invisible details. In our experiment, the threshold of block division is 0.3, and the size of $\mathcal{Q}$ is 10.

\subsection{Occupancy Inference}
Our occupancy inference network is based on PIFu \cite{saito2019pifu} which consists of a Stacked Hourglass network \cite{newell2016stacked} as the image encoder and a MLP with 258, 1024, 512, 256, 128, and 1 neurons in each layer. To fully utilize the depth information, we also encode the depth image and add a PSDF feature to MLP. The PSDF of a 3D point $\mathbf{x}$ is defined as
\begin{equation}
    PSDF(\mathbf{x}) = \left\{\begin{matrix}
\mathbf{x}_z-d & ,\mathbf{x}_z-d < 0\\ 
0 & ,\mathbf{x}_z-d \geq 0
\end{matrix}\right.,
\end{equation}
where $\mathbf{x}_z$ is the z-axis coordinate of $\mathbf{x}$, and $d$ is the depth value sampled at the projected location of $\mathbf{x}$ on the depth image.
We render a synthetic dataset (\url{https://web.twindom.com/}) to generate depth and color images as training data, and utilize 5000 images to train this network. When training, the batch size is 4, the learning rate is $1\times 10^{-3}$ and the number of epochs is 100.

\begin{figure}[t]
    \centering
    \includegraphics[width=\linewidth]{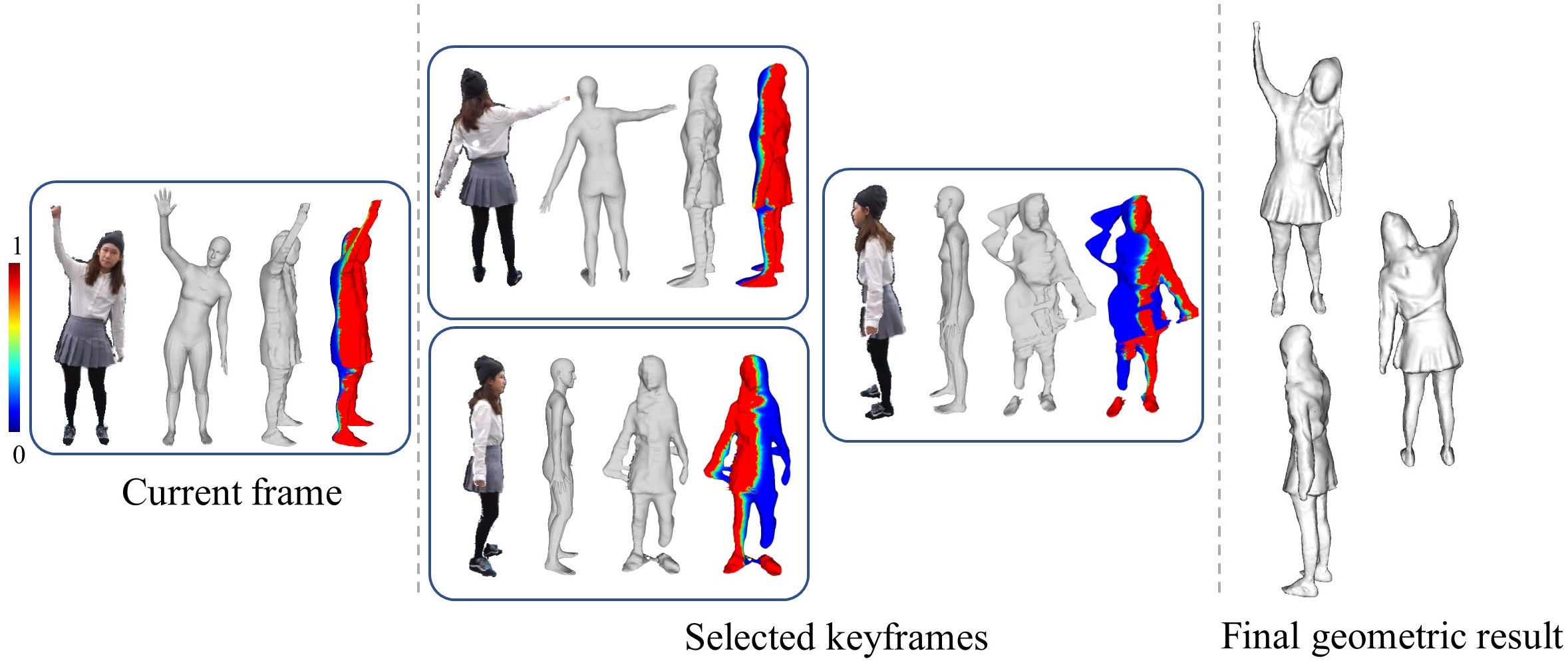}
    \caption{Intermediate results during reconstruction of one frame. In each rectangle, from left to right are the reference color image, warped SMPL, dense reconstruction and color-encoded blending weights, respectively.}
    \label{fig:intermediate results}
    \vspace{-0.4cm}
\end{figure}

\begin{figure}[t]
    \centering
    \includegraphics[width=\linewidth]{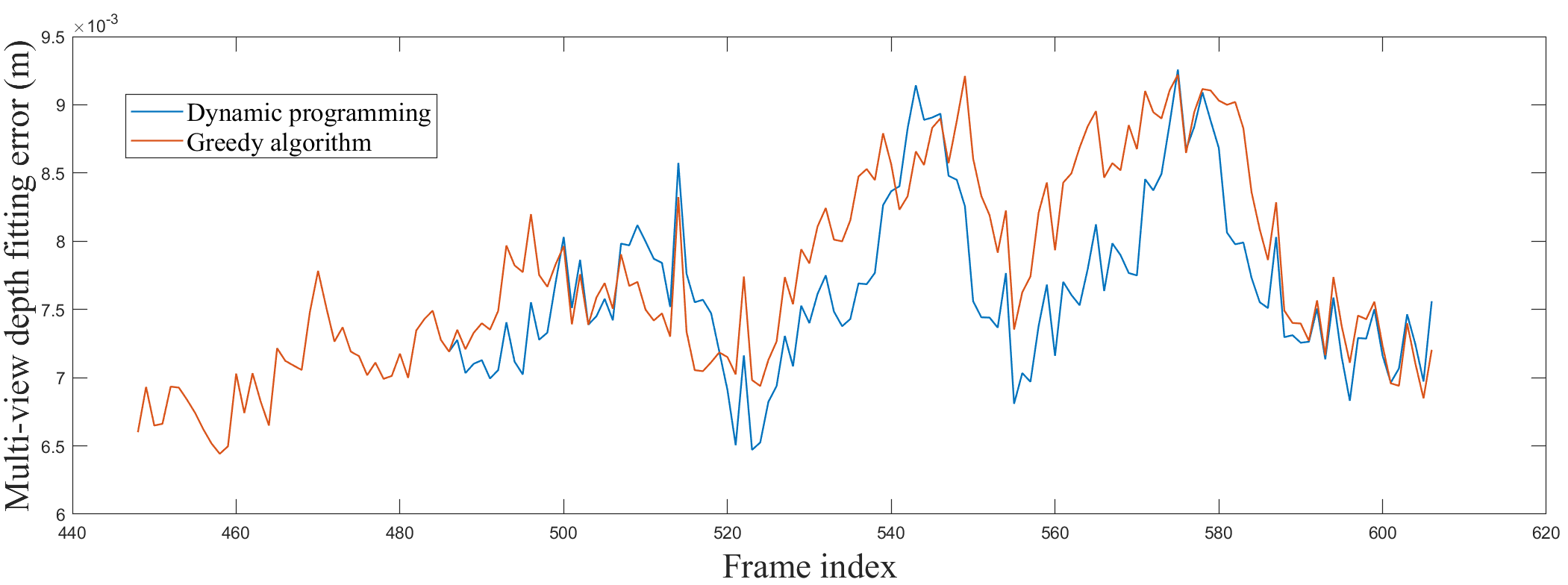}
    \caption{Quantitative comparison against the greedy algorithm on the mean multi-view depth fitting error.}
    \label{fig:quan greedy}
    \vspace{-0.4cm}
\end{figure}

\subsection{Collision Handling}
\noindent\textbf{Collision Detection} Based on the SMPL model, we first calculate the distance between every two vertices, and construct a symmetrical distance matrix $D\in\mathbb{R}^{N\times N}$ ($N$ is the vertex number), and $D_{ij}$ is the distance between the $i$-th and $j$-th vertices. For the current frame (the $t$-th frame) and one keyframe (the $k$-th frame), we calculate $D^{t}$ and $D^{k}$ respectively. And the collided vertices on the current SMPL are detected by the condition:
\begin{equation}
\begin{split}
\label{eq:collision condition}
    D_{ij}^t < \tau_1,\quad D_{ij}^k > \tau_2,
\end{split}
\end{equation}
which means that the $i$-th and $j$-th vertices collide with each other in the current frame. All the collided SMPL vertices are denoted as $\mathbf{V}_{\text{c}}$. The voxels corresponding to $\mathbf{V}_{\text{c}}$ are also detected. In our experiment, $\tau_1=0.02m$ and $\tau_2=0.05m$.

\noindent\textbf{Searching No-collision Frame} Starting from the current frame, we search for a no-collision frame both forward and backward. We calculate the proportion of SMPL vertices that belong to $\mathbf{V}_{\text{c}}$ but do not meet the collision condition (Eq.~\ref{eq:collision condition}) during searching each frame. If the proportion is less $30\%$, the no-collision frame is found. Then we deform the reconstructed model of the no-collision frame to the current frame, and integrate the deformed model into implicit surface fusion. In the fusion procedure, we first generate a 3D mask in the volume by the collision flag, and then perform 3D distance transform to generate a continuous weight volume for integrating the deformed model.

\subsection{Parameter Setting}
In the keyframe selection, we select $K=4$ keyframes for each frame. And in the first iteration, $\lambda_{\text{visibility}} = 1.5$, and $\lambda_{\text{visibility}} = 3.0$ in the subsequent iterations. In the adaptive blending weight, we set $\tau=0.02$ and $\sigma=100$.

\begin{figure}[t]
    \centering
    \includegraphics[width=\linewidth]{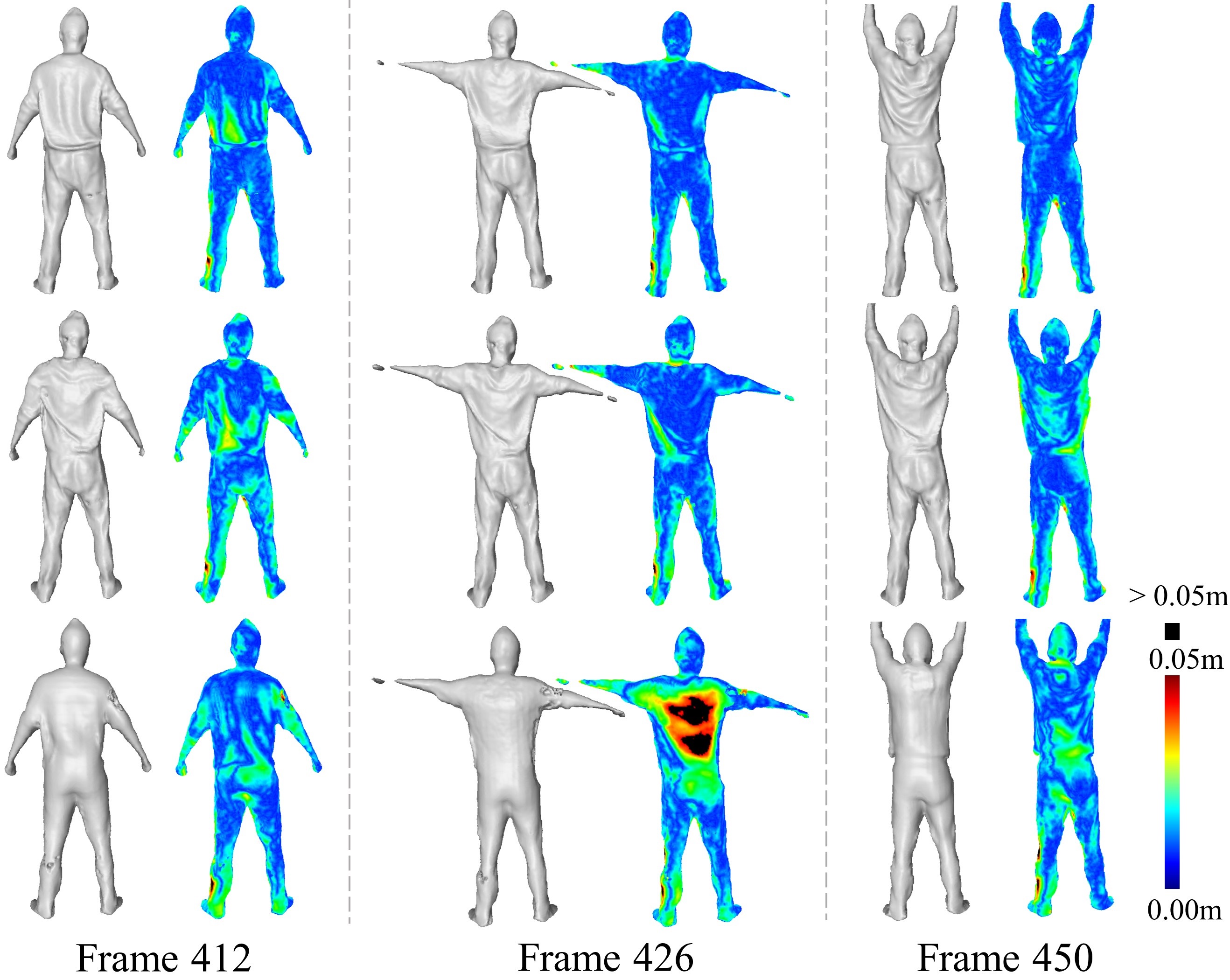}
    \caption{Reconstructed invisible details and visualization of per-vertex errors in several frames. From top to bottom are results with both energies, without the pose energy and without the visibility energy, respectively.}
    \label{fig: supp quantitative ablation study kf selection}
    \vspace{-0.4cm}
\end{figure}

\section{Performance}
We implement POSEFusion on one PC with one NVIDIA RTX 2080Ti, and the performance mainly depends on the number of keyframes. The time of processing one keyframe is almost 1.87 secs, which consists of 1.53 secs for deforming points and collision detection, 0.02 secs for loading images, 0.3 secs for occupancy inference, and 0.02 secs for blending occupancy values. In our experiment, we perform the pose-guided keyframe selection to select $K=4$ keyframes, and consider the current frame and the 3 frames around it as keyframes as well for denoising. So the keyframe number is $7$, and the time cost for reconstructing one frame is 13 secs. Moreover, we believe the runtime of this system could be further improved by TensorRT and other GPU toolkits.

\section{Additional Experiments}

\noindent\textbf{Intermediate Results}
In Fig.~\ref{fig:intermediate results}, given one current frame and its keyframe set, we demonstrate all the intermediate results including the reference color image, warped SMPL, dense reconstruction and color-encoded blending weights within the current frame and each keyframe.

\noindent\textbf{Quantitative Comparison against Greedy Algorithm} Fig.~\ref{fig:quan greedy} demonstrates the mean multi-view fitting errors of the results using the dynamic programming (DP) and greedy algorithm, respectively. And the average error of the whole sequence using DP and greedy algorithm are 0.7507 cm and 0.7732 cm, respectively. It shows that dynamic programming can produce the global optimum by considering the energy sum of the whole sequence, and achieve more physically accurate reconstruction.

\noindent\textbf{Quantitative Ablation Study of Pose-guided Keyframe Selection} Besides the mean vertex errors demonstrated in the main paper 
(Fig.~\ref{fig:quantitative ablation study kf selection})
, we also show the reconstructed results and visualization of per-vertex errors in this ablation study in Fig.~\ref{fig: supp quantitative ablation study kf selection}. It shows that with both visibility and pose guidance, the selected keyframe can cover invisible regions and share the similar poses with the current frame, so that physically plausible invisible details are recovered.

\clearpage
{\small
\bibliographystyle{ieee_fullname}
\bibliography{egbib}
}

\end{document}